\pdfoutput=1

\documentclass[11pt]{article}

\usepackage{acl}

\usepackage{times}
\usepackage{latexsym}

\usepackage[T1]{fontenc}

\usepackage[utf8]{inputenc}

\usepackage{microtype}

\usepackage{graphicx}
\usepackage{amsmath}
\usepackage{booktabs}
\usepackage{nicefrac}
\usepackage{enumitem}
\usepackage{sidecap}
\usepackage{multirow}
\usepackage{marvosym}
\usepackage{amsmath,graphicx}
\usepackage{amssymb}
\usepackage{amsthm}
\usepackage{mathtools}
\usepackage{bm}
\usepackage{bbm}

\title{Efficient Machine Translation Domain Adaptation}

\author{Pedro Henrique Martins\textsuperscript{\Neptune} \quad
        Zita Marinho\textsuperscript{\Moon\Scorpio} \quad
        Andr\'e F.~T. Martins\textsuperscript{\Neptune\Pluto\Saturn} \\
\textsuperscript{\Neptune}Instituto de Telecomunica\c{c}\~oes~
\textsuperscript{\Moon}DeepMind~
\textsuperscript{\Scorpio}Institute of Systems and Robotics\\
\textsuperscript{\Pluto}LUMLIS (Lisbon ELLIS Unit), Instituto Superior T\'ecnico~
\textsuperscript{\Saturn}Unbabel\\
Lisbon, Portugal\\
\href{mailto:pedrohenriqueamartins@tecnico.ulisboa.pt}{\tt pedrohenriqueamartins@tecnico.ulisboa.pt},\\
\href{mailto:zmarinho@google.com}{\tt zmarinho@google.com}, \quad
\href{mailto:andre.martins@unbabel.com}{\tt andre.t.martins@tecnico.ulisboa.pt}.
}

\begin{document}
\maketitle
\begin{abstract}
Machine translation models struggle when translating out-of-domain text, which makes domain adaptation a topic of critical importance. However, most domain adaptation methods focus on fine-tuning or training the entire or part of the model on every new domain, which can be costly. On the other hand, semi-parametric models have been shown to successfully perform domain adaptation by retrieving examples from an in-domain datastore \citep{khandelwal2020nearest}. 
A drawback of these retrieval-augmented models, however, is that they tend to be substantially slower. In this paper, we explore several approaches to speed up nearest neighbor machine translation. We adapt the methods recently proposed by \citet{he2021efficient} for language modeling, and introduce a simple but effective caching strategy that avoids performing retrieval when similar contexts have been seen before. Translation quality and runtimes for several domains show the effectiveness of the proposed solutions.\footnote{The code is available at \url{https://github.com/deep-spin/efficient_kNN_MT}.}
\end{abstract}

\section{Introduction}
Modern neural machine translation models are mostly parametric \citep{bahdanau2015neural, vaswani2017attention}, meaning that, for each input, the output  depends only on a fixed number of model parameters, obtained using some training data, hopefully in the same domain. 
However, when running machine translation systems in the wild, it is often the case that the model is given input sentences or documents from domains that were not part of the training data, which frequently leads to subpar translations. One solution is training or fine-tuning the entire model or just part of it for each domain, but this can be expensive and may lead to catastrophic forgetting \citep{saunders2021domain}.

Recently, an approach that has  achieved promising results is augmenting parametric models with a retrieval component, leading to \textit{semi-parametric} models  \citep{gu2018search, zhang2018guiding, bapna2019non, khandelwal2020nearest, meng2021fast, zheng2021adaptive, jiang2021learning}. These models  construct a datastore based on a set of source / target sentences or word-level contexts (translation memories) and retrieve similar examples from this datastore, using this information in the generation process. This allows having only one model that can be used for every domain. However, the model's runtime increases with the size of the domain's datastore and searching for related examples on large datastores can be  computationally very expensive: for example, when retrieving $64$ neighbors from the datastore, the model may become two orders of magnitude slower \citep{khandelwal2020nearest}. 
Due to this, some recent works have proposed methods that aim to make this process more efficient.
\citet{meng2021fast} proposed constructing a different datastore for each source sentence, by first searching for the neighbors of the source tokens; and  \citet{he2021efficient} proposed several techniques -- datastore pruning, adaptive retrieval, dimension reduction -- for nearest neighbor language modeling.

In this paper, we adapt several methods proposed by \citet{he2021efficient} to machine translation, and we further propose a new approach that increases the model's efficiency: the use of a retrieval distributions cache.
By caching the $k$NN probability distributions, together with the corresponding decoder representations, for the previous steps of the generation of the current translation(s), the model can quickly retrieve the retrieval distribution when the current representation is similar to a cached one, instead of having to search for neighbors in the datastore at every single step.

We perform a thorough analysis of the model's efficiency on a controlled setting, which shows that the combination of our proposed techniques results in a model, the efficient $k$NN-MT, which is approximately twice as fast as the vanilla $k$NN-MT. This comes without harming translation performance, which is, on average, more than $8$ BLEU points and $5$ COMET points better than the base MT model. 

In sum, this paper presents the following contributions:
\begin{itemize}
    \item We adapt the methods proposed by \citet{he2021efficient} for efficient nearest neighbor language modeling to machine translation.
    \item We propose a caching strategy to store the retrieval probability distributions, improving the translation speed.
    \item We compare the efficiency and translation quality of the different methods, which show the benefits of the proposed and adapted techniques.
\end{itemize}

\section{Background}
When performing machine translation, the model is given a source sentence or document, $\bm{x}=[x_1, \dots, x_L]$, on one language, and the goal is to output a translation of the sentence in the desired language, $\bm{y}=[y_1, \dots, y_N]$. 
This is usually done using a parametric sequence-to-sequence model \citep{bahdanau2015neural,vaswani2017attention}, in which the encoder receives the source sentence as input and outputs a set of hidden states. Then, at each step $t$, the decoder attends to these hidden states and outputs a probability distribution $p_{\mathrm{NMT}}(y_t|\bm{y}_{<t}, \bm{x})$ over the vocabulary. Finally, these probability distributions are used to predict the output tokens, typically with beam search.

\subsection{Nearest Neighbor Machine Translation}
\citet{khandelwal2020nearest} introduced a nearest neighbor machine translation model, $k$NN-MT, which is a semi-parametric model. This means that besides having a parametric component that outputs a probability distribution over the vocabulary, $p_{\mathrm{NMT}}(y_t|\bm{y}_{<t}, \bm{x})$, the model also has a nearest neighbor retrieval mechanism, which allows direct access to a datastore of examples.

More specifically, we build  a datastore $\mathcal{D}$ which consists of a key-value memory, where each entry key is the decoder's output representation, $\bm{f}(\bm{x},\bm{y}_{<t})$, and the value is the target token $y_t$:
\begin{equation}
    \mathcal{D} \!=\! \left\{\left(\bm{f}(\bm{x},\bm{y}_{<t}\right),y_t) \; \forall y_t \! \in  \bm{y} \mid (\bm{x},\bm{y})\! \in \!\left(\mathcal{X}, \mathcal{Y}\right)\right\}\!,
\end{equation}
where $\left(\mathcal{X}, \mathcal{Y}\right)$ corresponds to a set of parallel source and target sequences.
Then, at inference time, the model searches the datastore to retrieve the set of $k$ nearest neighbors $\mathcal{N}$. Using their distances $d(\cdot)$ to the current decoder's output representation, we can compute the retrieval distribution
$p_{k\mathrm{NN}}(y_t|\bm{y}_{<t}, \bm{x})$ as:
\begin{align}
    &p_{k\mathrm{NN}}(y_t|\bm{y}_{<t}, \bm{x}) =\\ 
    &\dfrac{\sum_{(\bm{k}_j,v_j)\in \mathcal{N}}\mathbbm{1}_{y_t=v_j} \exp \left(-d\left(\bm{k}_j,\bm{f}(\bm{x},\bm{y}_{<t})\right)/T\right)}{\sum_{(\bm{k}_j,v_j)\in \mathcal{N}} \; \exp \left(-d\left(\bm{k}_j,\bm{f}(\bm{x},\bm{y}_{<t})\right)/T\right)}, \nonumber
\end{align}
where $T$ is the softmax temperature, $k_j$ denotes the key of the $j^{th}$ neighbor and $v_j$ its value. 
Finally, $p_{\mathrm{NMT}}(y_t|\bm{y}_{<t}, \bm{x})$ and $p_{k\mathrm{NN}}(y_t|\bm{y}_{<t}, \bm{x})$ are combined to obtain the final distribution, which is used to generate the translation through beam search, by performing interpolation:
\begin{align}
    p(y_t|\bm{y}_{<t}, \bm{x})= &(1-\lambda) \; p_{\mathrm{NMT}}(y_t|\bm{y}_{<t}, \bm{x}) \\
    &+ \lambda \; p_{k\mathrm{NN}}(y_t|\bm{y}_{<t},\bm{x}), \nonumber
\end{align}
where $\lambda$ is a hyper-parameter that controls the weights given to the two distributions. 

\section{Efficient $k$NN-MT}
\label{sec:efficient_knnmt}
In this section, we describe the approaches introduced by \citet{he2021efficient} to speed-up the inference time for nearest neighbor language modeling, such as pruning the datastore (\S \ref{sec:pruning}) and reducing the representations dimension (\S \ref{sec:pca}), which we adapt to machine translation.
We further describe a novel method that allows the model to have access to examples without having to search them in the datastore at every step, by maintaining a cache of the past retrieval distributions, for the current translation(s) (\S \ref{sec:cache}).

\subsection{Datastore Pruning}
\label{sec:pruning}
The goal of datastore pruning is to reduce the size of the datastore, so that the model is able to search for the nearest neighbors faster, without severely compromising the translation performance. 
To do so, we follow \citet{he2021efficient}, and use greedy merging. 
In greedy merging, we aim to merge datastore entries that share the same value (target token) while their keys are close to each other in vector space. To do this, we first need to find the $k$ nearest neighbors of every entry of the datastore, where $k$ is a hyper-parameter. Then, if in the set of neighbors, retrieved for a given entry, there is an entry which has not been merged before and has the same value, we merge the two entries, by simply removing the neighboring one.

\subsection{Dimension Reduction}
\label{sec:pca}
The decoder's output representations, $\bm{f}(\bm{x},\bm{y}_{<t})$ are, usually, high-dimensional (1024, in our case). This leads to a high computational cost when computing vector distances, which are needed for retrieving neighbors from the datastore. To alleviate this, we follow \citet{he2021efficient}, and use principal component analysis (PCA), an efficient dimension reduction method, to reduce the dimension of the decoder's output representation to a pre-defined dimension, $d$, and generate a compressed datastore.

\subsection{Cache}
\label{sec:cache}
The model does not need to search the datastore at every step of the translation generation in order to do it correctly. Here, we aim to predict when it needs to retrieve neighbors from the datastore, so that, by only searching the datastore in the necessary steps, we can increase the generation speed.

\paragraph{Adaptive retrieval. } 
To do so, first we follow \citet{he2021efficient}, and use a simple MLP to predict the value of the interpolation coefficient $\lambda$ at each step. Then, we define a threshold, $\alpha$, so that the model only performs retrieval when $\lambda>\alpha$.
However, we observed that this leads to results (\S \ref{sec:results_retrieval}) similar to randomly selecting when to search the datastore. 
We posit that this occurs because it is difficult to predict when the model should perform retrieval, for domain adaptation \citep{he2021efficient}, and because in machine translation error propagation occurs more prominently than in language modeling.

\paragraph{Cache. } 
Because it is common to have similar contexts along the generation process, when using beam search, the model can be often retrieving similar neighbors at different steps, which is not efficient. 
To avoid repeating searches on the datastore for similar context vectors, $\bm{f}(\bm{x},\bm{y}_{<t})$, we propose keeping a cache of the previous retrieval distributions, of the current translation(s). 
More specifically, at each step of the generation of $\bm{y}$, we add the decoder's representation vector along with the retrieval distribution $p_{k\mathrm{NN}}(y_t|\bm{y}_{<t}, \bm{x})$, corresponding to all beams, $\mathcal{B}$, to the cache $\mathcal{C}$:
\begin{equation}
\mathcal{C} \!\!=\!\! \left\{\left(\bm{f}(\bm{x},\bm{y}_{<t}),p_{k\mathrm{NN}}(y_t|\bm{y}_{<t},\! \bm{x})\right) \! \forall  y_t \! \in  \bm{y} \! \mid \! \bm{y}\! \in \!\mathcal{B}\right\}\!.
\end{equation}
Then, at each step of the generation, we compute the Euclidean distance between the current decoder's representation and the  keys on the cache.
If all distances are bigger than a threshold $\tau$, the model searches the datastore to find the nearest neighbors. Otherwise, the model retrieves, from the cache, the retrieval distribution that corresponds to the closest key.

\section{Experiments}
\paragraph{Dataset and metrics. } We perform experiments on the Medical, Law, IT, and Koran domain data of the multi-domains dataset \citep{koehn2017six} re-splitted by \citet{aharoni2020unsupervised}. To build the datastores we use the in-domain training sets which have from 17,982 to 467,309 sentences. The validation and test sets have 2,000 sentences. 

To evaluate the models we use BLEU \citep{papineni2002bleu,post2018call} and COMET \citep{rei2020comet}. 

\paragraph{Settings.} We use the WMT'19 German-English news translation task winner \citep{ng2019facebook} (with 269 M parameters), available on the Fairseq library \cite{ott2019fairseq}, as the base MT model.

As baselines, we consider the base MT model, the vanilla $k$NN-MT model \citep{khandelwal2020nearest}, and the Fast $k$NN-MT model \citep{meng2021fast}. For all models, which perform retrieval, we select the hyper-parameters, for each method and each domain, by performing grid search on $k \in \{8,16,32,64\}$ and $\lambda \in \{0.5,0.6,0.7,0.8\}$. The selected values are stated in Table \ref{table:hyperparameters} of App. \ref{sec:hyperparameters}.

For the vanilla $k$NN-MT model and the efficient $k$NN-MT we follow \citet{khandelwal2020nearest} and use the Euclidean distance to perform retrieval and the proposed softmax temperature. For the Fast $k$NN-MT, we use the cosine distance and the softmax temperature proposed by \citet{meng2021fast}. 
For the efficient $k$NN-MT we selected parameters that ensure a good speed/quality trade-off: $k=2$ for datastore pruning, $d=256$ for PCA, and $\tau=6$ as the cache threshold. Results for each methods using different parameters are reported in App. \ref{sec:additional_results}.

\subsection{Results}
\label{sec:results}

\begin{table*}[t]
\centering \small
\setlength{\tabcolsep}{1.3ex}
\begin{tabular}
{lccccc@{\hspace{4ex}}ccccc}
\toprule
& \multicolumn{5}{c}{BLEU} & \multicolumn{5}{c}{COMET} \\
& Medical & Law & IT & Koran & Average & Medical & Law & IT & Koran & Average \\
\midrule
\textbf{Baselines} \\
Base MT & 40.01  & 45.64 & 37.91 & 16.35 & 34.98 & .4702 & .5770 & .3942 & -.0097 & .3579 \\
$k$NN-MT & 54.47 & 61.23 & 45.96 & 21.02 & 45.67 & .5760 & .6781 & .5163 & .0480 & .4546 \\
Fast $k$NN-MT & 52.90 & 55.71 & 44.73 & 21.29 & 43.66 & .5293 & .5944 & .5445 & -.0455 & .4057\\
\midrule
\textbf{Efficient $k$NN-MT} \\
cache &  53.30 & 59.12 & 45.39 & 20.67 & 44.62 & .5625 & .6403 & .5085 & .0346 & .4365 \\
PCA + cache & 53.58 & 58.57 & 46.29 & 20.67 & 44.78 & .5457 & .6379 & .5311 & -.0021 & .4282 \\
PCA + pruning & 53.23 & 60.38 & 45.16 & 20.52 & 44.82 & .5658 & .6639 & .4981 & .0298 & .4394 \\
PCA + cache + pruning & 51.90 & 57.82 & 44.44 & 20.11 & 43.57 & .5513 & .6260 & .4909 & -.0052 & .4158 \\
\bottomrule
\end{tabular}
\caption{BLEU and COMET scores on the multi-domains test set, for a batch size of 8. }
\label{table:results}
\end{table*}

\begin{figure*}[t]
    \centering
    \includegraphics[width=4.06cm]{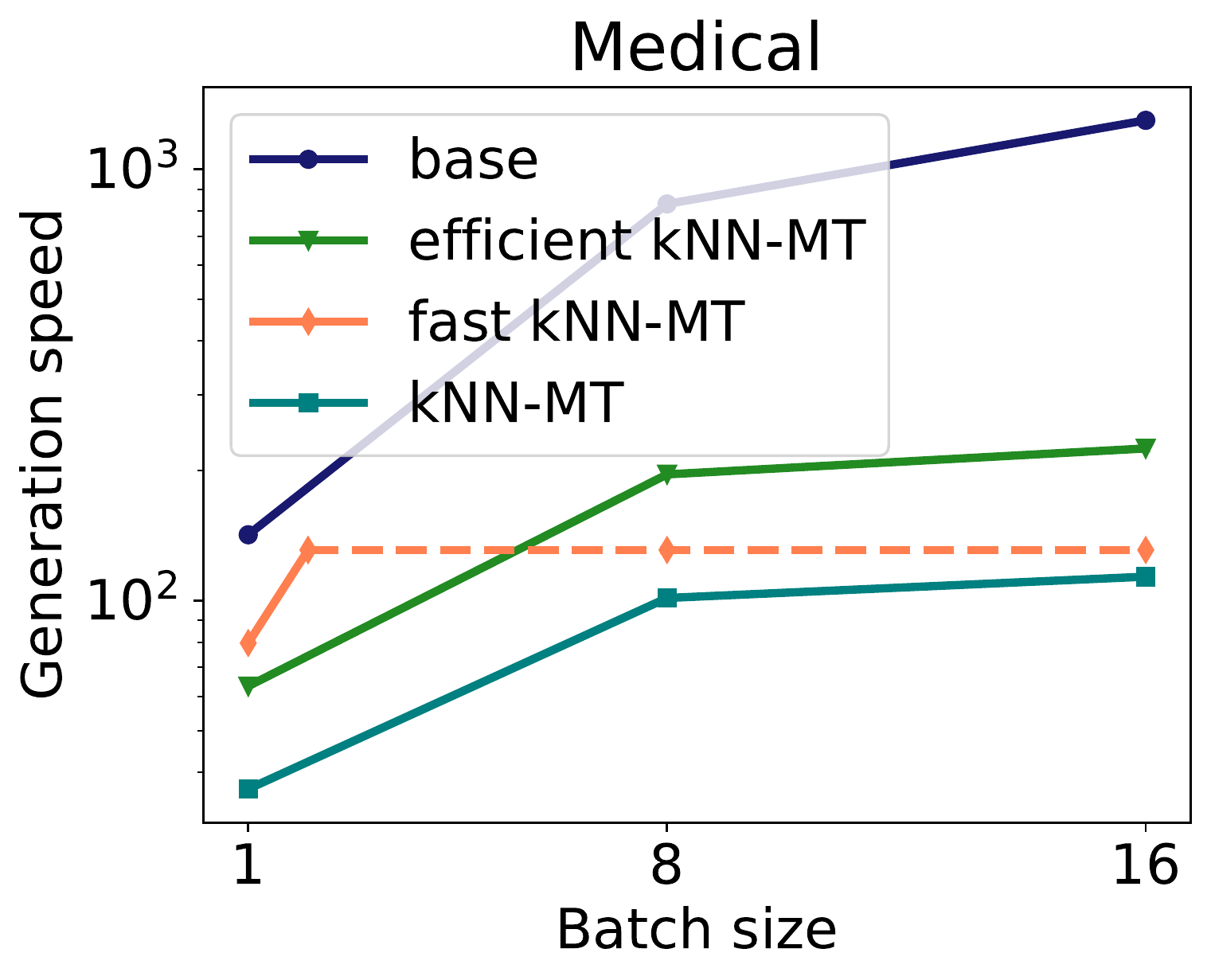}
    \includegraphics[width=3.88cm]{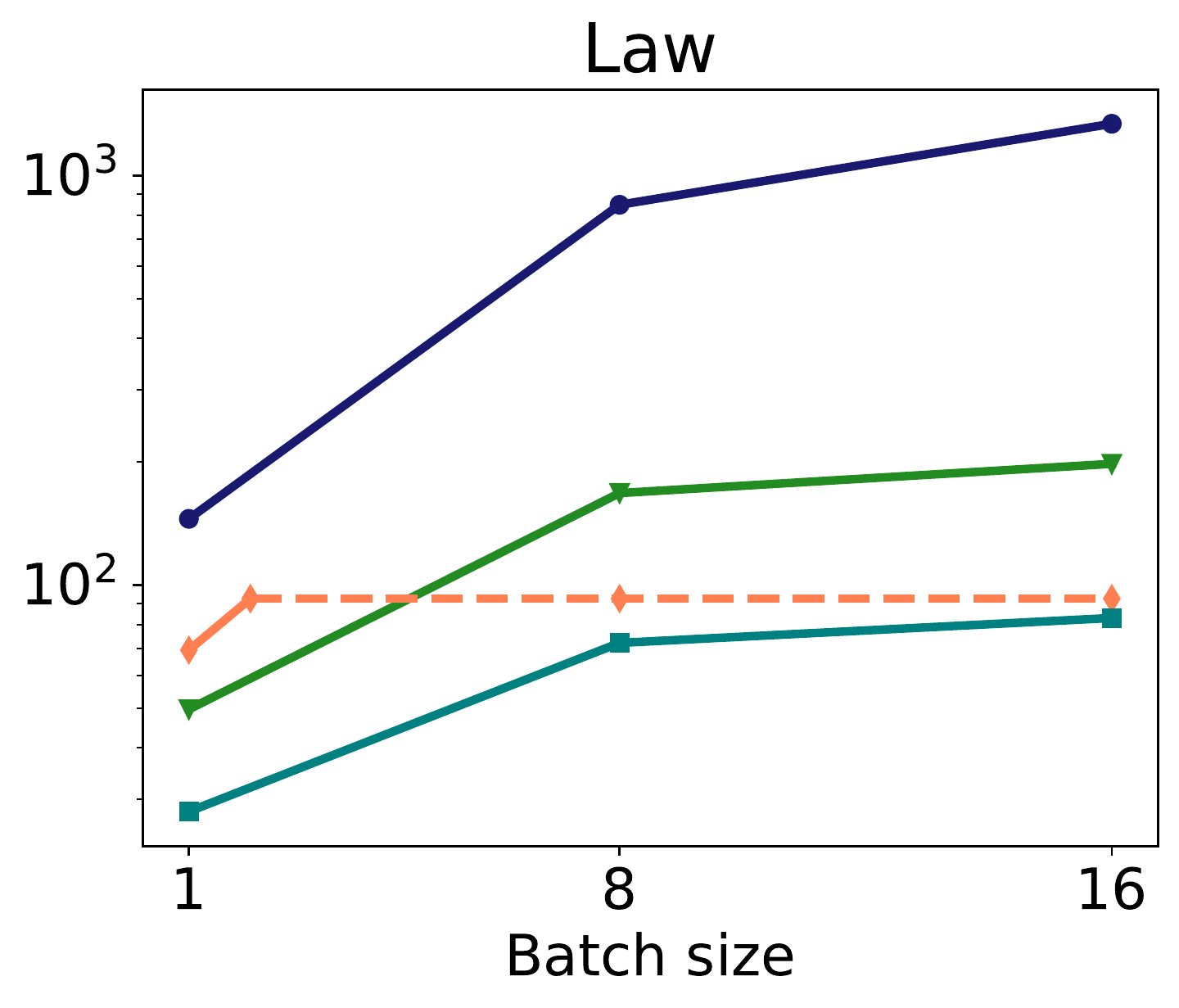}
    \includegraphics[width=3.88cm]{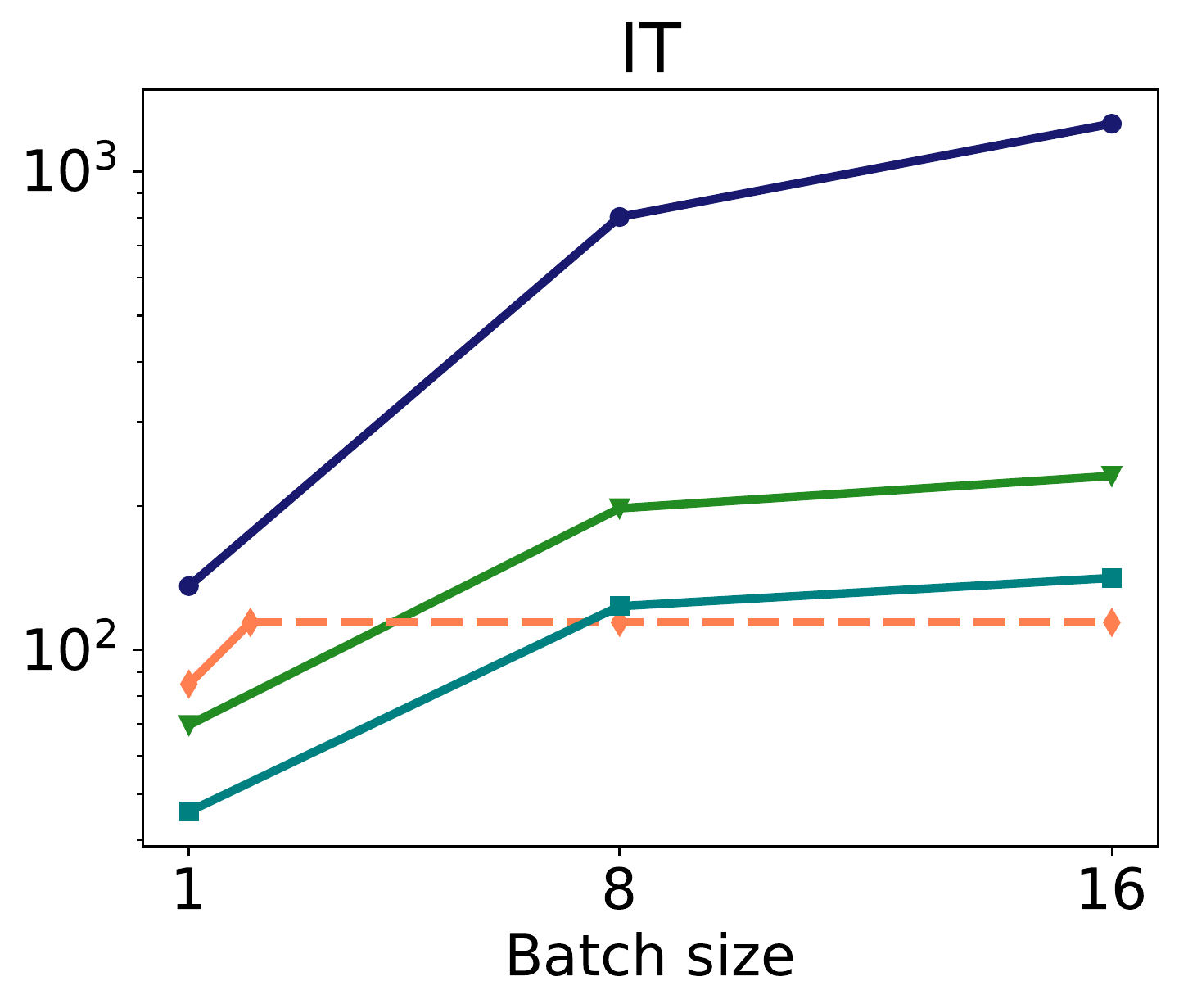}
    \includegraphics[width=3.88cm]{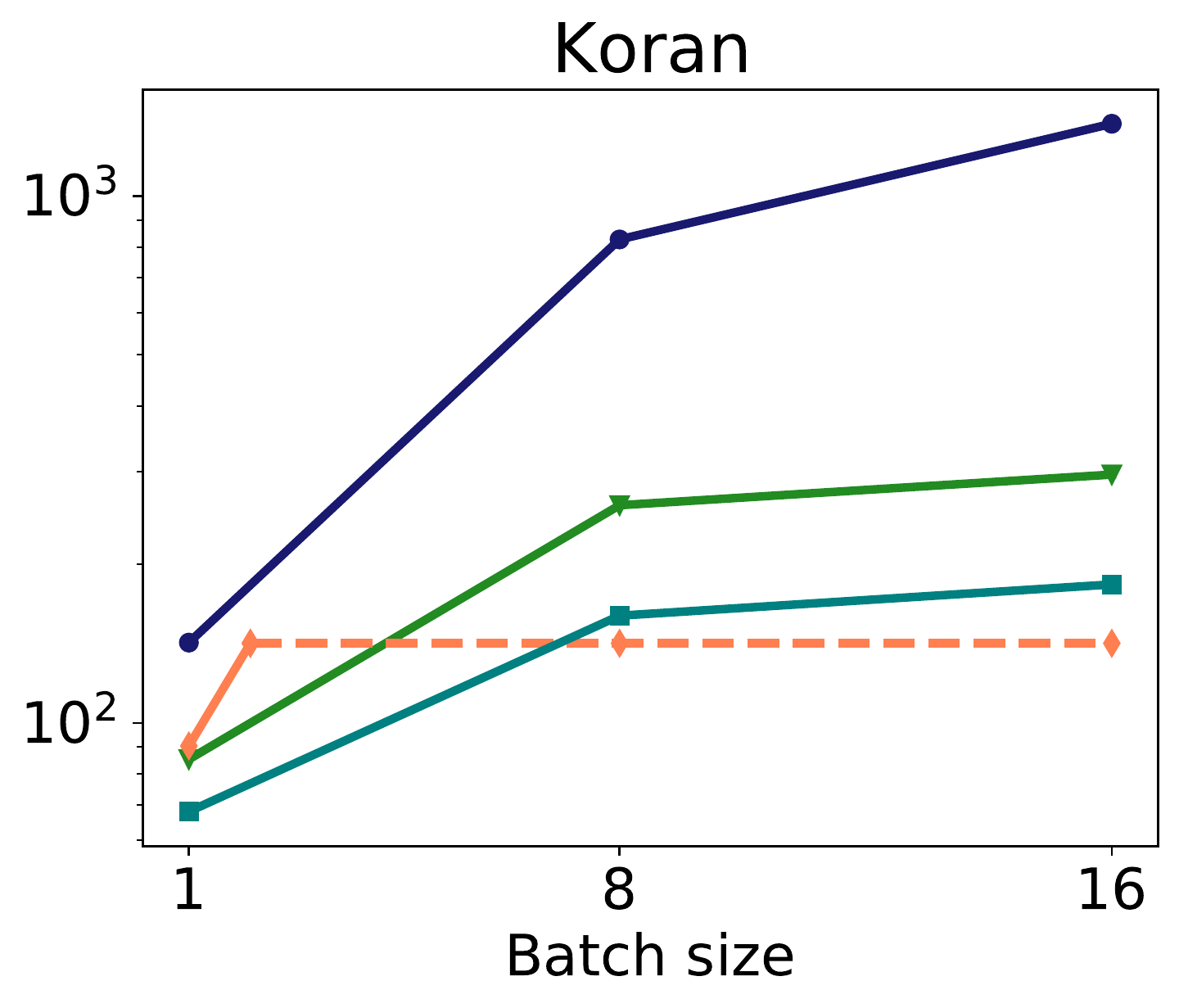}
    \caption{Plots of the generation speed (tokens/s) for the different models on the medical, law, IT, and Koran domains, for different batch sizes (1,8,16). The generation speed (y-axis) is in log scale. When using the Fast $k$NN-MT model, the maximum batch size that we are able to use is 2, due to out of memory errors.}
    \label{fig:generation_speed}
\end{figure*}

The translation scores are reported on Table~\ref{table:results}. We can clearly see that both Fast $k$NN-MT and the efficient $k$NN-MT (combining the different methods) do not hurt the translation performance substantially, still leading to, on average, $8$ BLEU points and $5$ COMET points more than the base MT model.

\subsection{Generation speed}
\label{sec:generation_speed}

\paragraph{Computational infrastructure. }
All experiments were performed on a server with 3 RTX 2080 Ti (11 GB), 12 AMD Ryzen 2920X CPUs (24 cores), and 128 Gb of RAM. For the generation speed measurements, we ran each model on a single GPU while no other process was running on the server, to have a controlled environment. To search the datastore, we used the FAISS library \citep{johnson2019billion}. When using the vanilla $k$NN-MT and efficient $k$NN-MT, the nearest neighbor search is performed on the CPUs, since not all datastores fit into memory, while when using the Fast $k$NN-MT this is done on the GPU.

\paragraph{Analysis. }
As can be seen on the plots of Figure~\ref{fig:generation_speed}, for a batch size of 1 Fast $k$NN-MT leads to a generation speed higher than our proposed method and vanilla $k$NN-MT. However, because of its high memory requirements, we are not able to run Fast $k$NN-MT for batch sizes larger than 2, on the computational infrastructure stated above. 
On the contrary, when using the proposed methods (efficient $k$NN-MT) we are able to run the model with higher batch sizes, achieving superior generation speeds to Fast $k$NN-MT and vanilla $k$NN-MT, and reducing the gap to the base MT model.

\paragraph{Ablation. }
\begin{figure}[ht]
    \centering
    \includegraphics[width=\columnwidth]{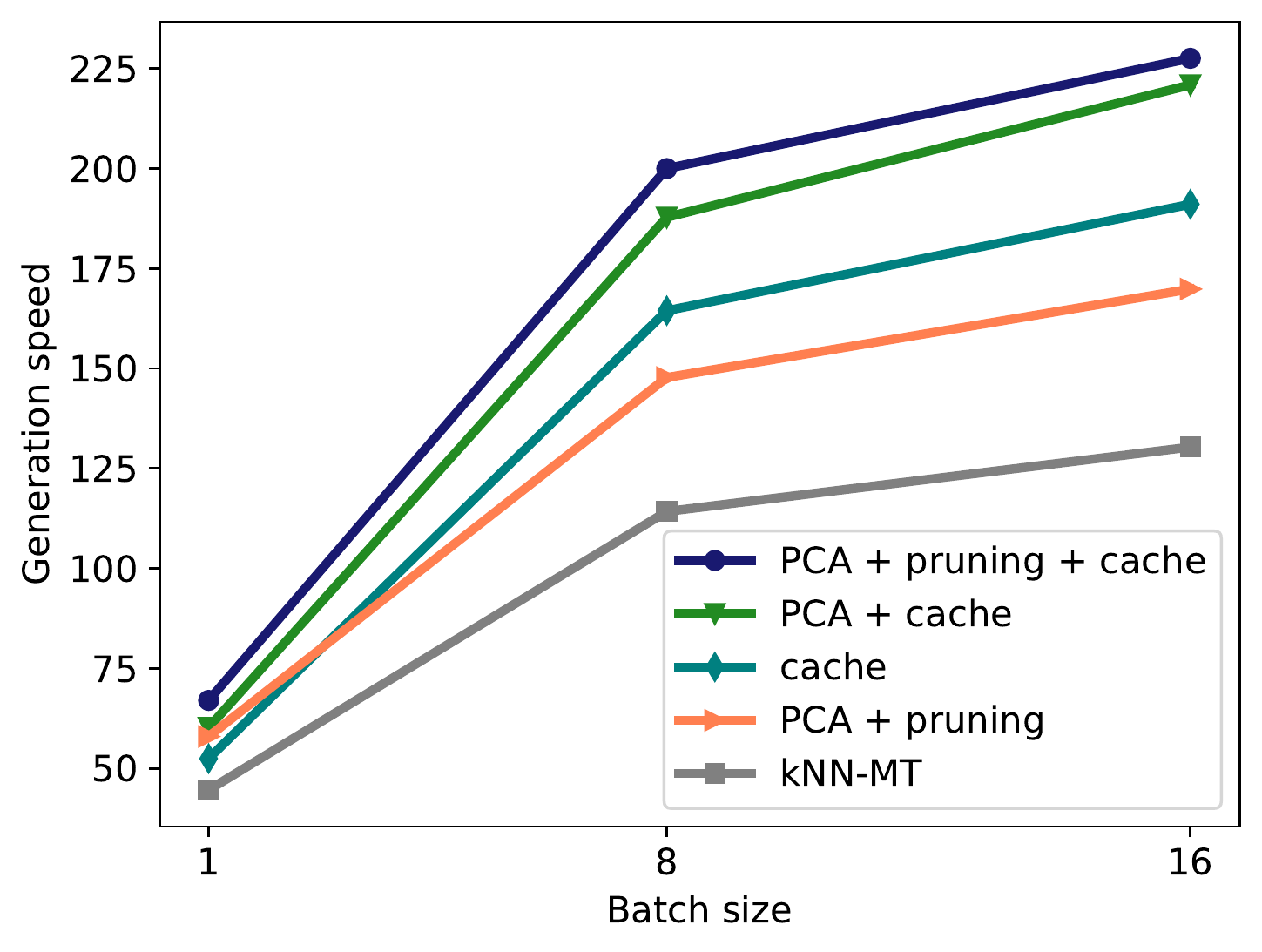}
    \caption{Plot of the generation speed (tokens/s), averaged across domains, for different combinations of the proposed methods.}
    \label{fig:generation_speed_ablation}
\end{figure}

We plot the generation speed for different combinations of the proposed methods (averaged across domains), for several batch sizes, on Figure~\ref{fig:generation_speed_ablation}. On this plot, we can clearly see that every method contributes to the speed-up achieved by the model that combines all approaches. Moreover, we can observe that the method which leads to the largest speed-up is the use of a cache of retrieval distributions, by saving, on average 57\% of the retrieval searches.

\section{Conclusion}
In this paper we propose the efficient $k$NN-MT, in which we combine several methods to improve the $k$NN-MT generation speed. First, we adapted to machine translation methods  that improve retrieval efficiency in language modeling \citep{he2021efficient}. Then we proposed a new method which consists on keeping in cache the previous retrieval distributions so that the model does not need to search for neighbors in the datastore at every step. Through experiments on domain adaptation, we show that the combination of the proposed methods leads to a considerable speed-up (up to 2x) without harming the translation performance substantially.

\section*{Acknowledgments}
This work was supported by the European Research Council (ERC StG DeepSPIN 758969), 
by the P2020 project MAIA (contract 045909), by the Funda\c{c}\~ao para a Ci\^encia e Tecnologia through project PTDC/CCI-INF/4703/2021 (PRELUNA, contract UIDB/50008/2020), and by contract PD/BD/150633/2020 in the scope of the  Doctoral Program  FCT - PD/00140/2013 NETSyS. We thank Junxian He, Graham Neubig, the SARDINE team members, and the reviewers for helpful discussion and feedback.

\bibliography{anthology,custom}
\bibliographystyle{acl_natbib}

\clearpage
\appendix

\section{Additional results}
\label{sec:additional_results}
In this section we report the BLEU scores as well as additional statistics for the different methods, when varying their hyper-parameters.

\subsection{Datastore pruning}
We report on Table \ref{table:results_pruning} the BLEU scores for datastore pruning, when varying the number of neighbors used for greedy merging, $k$. The resulting datastore sizes are presented on Table \ref{table:pruning_sizes}. 

\begin{table}[h!]
\centering \small
\setlength{\tabcolsep}{1.2ex}
\begin{tabular}
{lccccc}
\toprule
& Medical & Law & IT & Koran & Average \\
\midrule
$k$NN-MT & 54.47 & 61.23 & 45.96 & 21.02 & 45.67 \\
\midrule
$k=1$ & 53.60 & 60.23 & 45.03 & 20.81 & 44.92 \\
$k=2$ & 52.95 & 59.40 & 44.76 & 20.12 & 44.31 \\
$k=5$ & 51.63 & 57.55 & 44.07 & 19.29 & 43.14 \\
\bottomrule
\end{tabular}
\caption{BLEU scores on the multi-domains test set when performing datastore pruning with several values of $k$, for a batch size of 8. }
\label{table:results_pruning}
\end{table}

\begin{table}[h!]
\centering \small
\setlength{\tabcolsep}{1ex}
\begin{tabular}
{lcccc}
\toprule
& Medical & Law & IT & Koran  \\
\midrule
$k$NN-MT & 6,903,141 & 19,061,382 & 3,602,862 & 524,374 \\
\midrule
$k=1$ & 4,780,514 & 13,130,326 & 2,641,709 & 400,385 \\
$k=2$ & 4,039,432 & 11,103,775 & 2,303,808 & 353,007 \\
$k=5$ & 3,084,106 & 8,486,551 & 1,852,191 & 290,192 \\
\bottomrule
\end{tabular}
\caption{Sizes of the in-domain datastores when performing datastore pruning with several values of $k$, for a batch size of 8. }
\label{table:pruning_sizes}
\end{table}

\subsection{Dimension reduction}
We report on Table \ref{table:results_pca} the BLEU scores for dimension reduction, when varying the output dimension $d$. 

\begin{table}[h!]
\centering \small
\setlength{\tabcolsep}{1.2ex}
\begin{tabular}
{lccccc}
\toprule
& Medical & Law & IT & Koran & Average \\
\midrule
$k$NN-MT & 54.47 & 61.23 & 45.96 & 21.02 & 45.67 \\
\midrule
$d=512$ & 55.06 & 62.04 & 46.98 & 21.24 & 46.33 \\
$d=256$ & 54.52 & 61.84 & 46.68 & 21.57 & 46.15 \\
$d=128$ & 53.94 & 61.17 & 45.46 & 21.35 & 45.48 \\
\bottomrule
\end{tabular}
\caption{BLEU scores on the multi-domains test set when performing PCA with different dimension, $d$, values, for a batch size of 8. }
\label{table:results_pca}
\end{table}

\subsection{Adaptive retrieval}
\label{sec:results_retrieval}
We report on Table \ref{table:results_adaptive} the BLEU scores for adaptive retrieval, when varying the threshold $\alpha$. The percentage of times the model performs retrieval is stated on Table \ref{table:adaptive_searches}. 

\begin{table}[h!]
\centering \small
\setlength{\tabcolsep}{1.2ex}
\begin{tabular}
{lccccc}
\toprule
& Medical & Law & IT & Koran & Average \\
\midrule
$k$NN-MT & 54.47 & 61.23 & 45.96 & 21.02 & 45.67 \\
\midrule
$\alpha=0.25$ & 45.52 & 49.91 & 37.97 & 16.36 & 37.44 \\
$\alpha=0.5$ & 52.84 & 59.36 & 38.58 & 18.08 & 42.22 \\
$\alpha=0.75$ & 53.90 & 60.87 & 43.05 & 19.91 & 44.43 \\
\bottomrule
\end{tabular}
\caption{BLEU scores on the multi-domains test set when performing adaptive retrieval for different values of the threshold $\alpha$, for a batch size of 8.}
\label{table:results_adaptive}
\end{table}

\begin{table}[h!]
\centering \small
\setlength{\tabcolsep}{1.5ex}
\begin{tabular}
{lcccc}
\toprule
& Medical & Law & IT & Koran \\
\midrule
$k$NN-MT & 100\% & 100\% & 100\% & 100\% \\
\midrule
$\alpha=0.25$ & 78\% & 73\% & 38\% & 4\% \\
$\alpha=0.5$ & 96\% & 96\% & 60\% & 61\% \\
$\alpha=0.75$ & 98\% & 99\% & 92\% & 91\% \\
\bottomrule
\end{tabular}
\caption{Percentage of times the model searches for neighbors on the datastore when performing adaptive retrieval for different values of the threshold $\alpha$, for a batch size of 8. }
\label{table:adaptive_searches}
\end{table}

\subsection{Cache}
We report on Table \ref{table:results_cache} the BLEU scores for a model using a cache of the retrieval distributions, when varying the threshold $\tau$. The percentage of times the model performs retrieval is stated on Table \ref{table:cache_searches}. 

\begin{table}[h!]
\centering \small
\setlength{\tabcolsep}{1.2ex}
\begin{tabular}
{lccccc}
\toprule
& Medical & Law & IT & Koran & Average \\
\midrule
$k$NN-MT & 54.47 & 61.23 & 45.96 & 21.02 & 45.67 \\
\midrule
$\tau=2$ & 54.47 & 61.23 & 45.93 & 20.98 & 45.65 \\
$\tau=4$ & 54.17 & 61.10 & 46.07 & 21.00 & 45.58 \\
$\tau=6$ & 53.30 & 59.12 & 45.39 & 20.67 & 44.62 \\
$\tau=8$ & 30.06 & 23.01 & 25.53 & 16.08 & 23.67 \\
\bottomrule
\end{tabular}
\caption{BLEU scores on the multi-domains test set when using a retrieval distributions' cache for different values of the threshold $\tau$, for a batch size of 8.}
\label{table:results_cache}
\end{table}

\begin{table}[h!]
\centering \small
\setlength{\tabcolsep}{1.5ex}
\begin{tabular}
{lcccc}
\toprule
& Medical & Law & IT & Koran \\
\midrule
$k$NN-MT & 100\% & 100\% & 100\% & 100\% \\
\midrule
$\tau=2$ & 59\% & 51\% & 67\% & 64\% \\
$\tau=4$ & 50\% & 42\% & 57\% & 53\% \\
$\tau=6$ & 43\% & 35\% & 49\% & 45\% \\
$\tau=8$ & 26\% & 16\% & 29\% & 31\% \\
\bottomrule
\end{tabular}
\caption{Percentage of times the model searches for neighbors on the datastore when using a retrieval distributions' cache for different values of the threshold $\tau$, for a batch size of 8. }
\label{table:cache_searches}
\end{table}

\section{Hyper-parameters}
\label{sec:hyperparameters}
\begin{table*}[ht]
\centering \small
\setlength{\tabcolsep}{1.3ex}
\begin{tabular}
{lccc@{\hspace{5ex}}ccc@{\hspace{5ex}}ccc@{\hspace{5ex}}ccc}
\toprule
& \multicolumn{3}{c}{Medical} & \multicolumn{3}{c}{Law} & \multicolumn{3}{c}{IT} & \multicolumn{3}{c}{Koran} \\
& $k$ & $\lambda$ & $T$ & $k$ & $\lambda$ & $T$ & $k$ & $\lambda$ & $T$ & $k$ & $\lambda$ & $T$ \\
\midrule
$k$NN-MT & 8 & 0.7 & 10 & 8 & 0.8 & 10 & 8 & 0.7 & 10 & 8 & 0.6 & 100 \\
Fast $k$NN-MT & 16 & 0.7 & .015 & 32 & 0.6 & .015 & 8 & 0.6 & .02 & 16 & 0.6 & .05 \\
\midrule
cache & 8 & 0.7 & 10 & 8 & 0.8 & 10 & 8 & 0.7 & 10 & 8 & 0.6 & 100 \\
PCA + cache & 8 & 0.8 & 10 & 8 & 0.8 & 10 & 8  & 0.7 & 10 & 8  & 0.7 & 100 \\
PCA + pruning & 8 & 0.7 & 10 & 8 & 0.8 & 10 & 8 & 0.7 & 10 & 8 & 0.7 & 100 \\
PCA + cache + pruning & 8 & 0.7 & 10 & 8 & 0.8 & 10 & 8 & 0.7 & 10 & 8 & 0.7 & 100 \\
\bottomrule
\end{tabular}
\caption{Values of the  hyper-parameters: number of neighbors to be retrieved $k$, interpolation coefficient $\lambda$, and retrieval softmax temperature $T$.}
\label{table:hyperparameters}
\end{table*}
On Table \ref{table:hyperparameters} we report the values for the hyper-parameters: number of neighbors to be retrieved $k \in \{8,16,32,64\}$, the interpolation coefficient $\lambda \in \{0.5,0.6,0.7,0.8\}$, and retrieval softmax temperature $T$. For decoding we use beam search with a beam size of $5$.

\end{document}